\documentclass[10pt]{spie}  

\usepackage{amsfonts}
\usepackage{amsmath}
\usepackage{amssymb}
\usepackage{graphicx}
\usepackage[colorlinks=true, allcolors=black]{hyperref} 
\usepackage[utf8]{inputenc}
\usepackage{paralist} 
\usepackage{subfigure}
\usepackage{txfonts}  
\usepackage{units}

\newcommand{\reals}{\ensuremath{\mathrm{I\!R}}}

\newenvironment{myitemize}
  { \begin{compactitem} }
  { \end{compactitem} }

\title{Wearable Camera-Based Human Absolute Localization in Large Warehouses}

\author[a]{Ga\"el \'Ecorchard}
\author[a]{Karel Ko\v{s}nar}
\author[a]{Libor P\v{r}eu\v{c}il}
\affil[a]{Czech Institute of Informatics, Robotics, and Cybernetics, Czech Technical University in Prague, Czech Republic}

\authorinfo{Corresponding author: G. Écorchard, gael.ecorchard@cvut.cz.}

\begin{document}
\maketitle

\setlength{\abovedisplayshortskip}{2pt}
\setlength{\belowdisplayshortskip}{2pt}
\setlength{\abovedisplayskip}{0pt}
\setlength{\belowdisplayskip}{0pt}

\begin{abstract}
In a robotised warehouse, as in any place where robots move autonomously, a major issue is the localization or detection of human operators during their intervention in the work area of the robots.
This paper introduces a wearable human localization system for large warehouses, which utilize preinstalled infrastructure used for localization of automated guided vehicles (AGVs).
A monocular down-looking camera is detecting ground nodes, identifying them and computing the absolute position of the human to allow safe cooperation and coexistence of humans and AGVs in the same workspace.
A virtual safety area around the human operator is set up and any AGV in this area is immediately stopped.
 In order to avoid triggering an emergency stop because of the short distance between robots and human operators, the trajectories of the robots have to be modified so that they do not interfere with the human.
The purpose of this paper is to demonstrate an absolute visual localization method working in the challenging environment of an automated warehouse with low intensity of light, massively changing environment and using solely monocular camera placed on the human body.
\end{abstract}


\section{INTRODUCTION}

Modern automated warehouses and distribution centers are using automated guided vehicles (AGVs) to  achieve maximum efficiency, flexibility and agility.
A method for the management of such a robotised warehouse is to store items on shelves that are moved around by AGVs between storage space and so-called pick stations, where a human picks single items from the shelf and puts them directly into the box that will be sent to the end-customer.
As safety is a highest priority, the AGVs are confined in the storage space.

On the other hand, this solution has drawbacks as well.
Any human intervention, such as a maintenance operation or tidying a dropped item, results in a complete shutdown of the automated system before a human operator can enter the protected area.
Every intervention is then costly because all robots and not only the faulty one stop to be productive.
Also the pick stations can be placed only on the borders of the protected area.
It results in longer distances from the pick station to the rack with items.

For these reasons, there is a need to make the AGVs collaborative and allows the coexistence of humans and AGVs in the same operating space.
It will allow to make the maintenance during the operation and make the placing of the pick station optimal.

In order to achieve this, a virtual safety area around every human in the warehouse is set and the robot will stop as soon as the distance to the human operator is lower than the radius of the safety area.
To avoid triggering an emergency stop in normal operation, the trajectories of the robots have to be modified so that they do not interfere with the human.
As a consequence, the human operator needs to be localised in the warehouse.
It should be emphasized here that in the context of the current project the human localization system is not safety-related because the range-based safety system is an independent system that is not presented here.

The existing solutions of wearable human localization system the indoor environments make use of various sensors.
Some of them are based on a laser and SLAM, like the work of Mur-Artal et al.\cite{Mur-Artal15a}, or the one of Chen et al.\cite{chen2010indoor}, where the lasers are placed in a backpack.
Other solutions make use of an existing Wi-Fi infrastructure\cite{xu2015enhancing}.
Most often, one of more cameras are used for visual localization.
Murillo et al.\cite{murillo2012wearable} design a localization system using an omnidirectional camera placed on the person head like a hat.
Ming et al.\cite{li2018real} make use of a Tango smartphone for visual localization.

The majority of existing approaches rely on some form of a priori knowledge in form of a map, and expect, that the environment is more or less static.
This is not our case, as there are massive changes in the warehouse due to frequent movements of the racks with the items.
Moreover, racks being identical, warehouse environments offer strong visual aliasing that would confuse all methods that use a global map based on image features.

\begin{figure}[ht]
  \centering
  \includegraphics[width=4cm]{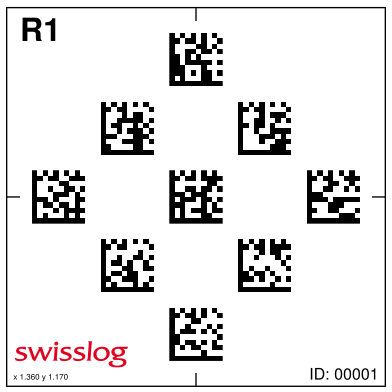}
  \caption{Example of ground node.}
  \label{fig:sticker}
\end{figure}

The localization of AGVs in the warehouse is often making use of a static pre-installed infrastructure.
In our case, unique ground nodes, shown in Fig.~\ref{fig:sticker}), are placed on the floor of the warehouse and their position is precisely measured in the absolute frame of reference.
Robots are able to detect these nodes and according to the identifier of the detected node determine their own absolute position and orientation.
Our approach takes advantage of using the existing infrastructure of the ground nodes.
Such markers cannot be found all over the warehouse with a maximum distance of approximately \unit[1.5]{m} between them.
However, for the reason that human operators cannot be required to pass over markers, this localization method must completed by a system providing localization when no marker is visible in the image, this is the role of the visual odometry.

The fusion of the visual odometry algorithm, which provides frequent and relative drifting position information, associated with a system that is able to provide the absolute position of the human operator in the warehouse with a down-facing monocular camera pointing at the ground nodes is used in the project.
The visual odometry algorithm is not detailed further in this paper.
The axis of the stereo camera for the visual odometry must be parallel to the ground and thus cannot be used by the marker-based localization.

The camera of the ground-node-based localization will be worn on the back of the operator in order not to interfere with his/her movements.
The main difference between analyzing the images with nodes on one side with the robots with constant small distance to the nodes, constant good light and smooth movements and, on the other side, with a human-worn camera are:

\begin{myitemize}
  \item Long distance to the node (camera on the lower part of the back; marker on the ground)
  \item Uncontrolled human movements
  \item Quick movements
  \item Low-light conditions
  \item Orientation relative to node has four solutions if the DataMatrix cannot be read
\end{myitemize}

\section{ALGORITHM}

The determination of the position occurs with the following steps:

\begin{myitemize}
  \item image undistortion;
  \item detection of the nodes in the image and extraction of a Region of Interest (ROI) around the best node;
  \item reading of DataMatrix codes in the ROI;
  \item computing of the camera's absolute position according to the position of the node in the warehouse and the position of its projection in the image.
\end{myitemize}

\subsection{Detection of the nodes}

\begin{figure}[ht]
  \centering
  \includegraphics[width=6cm]{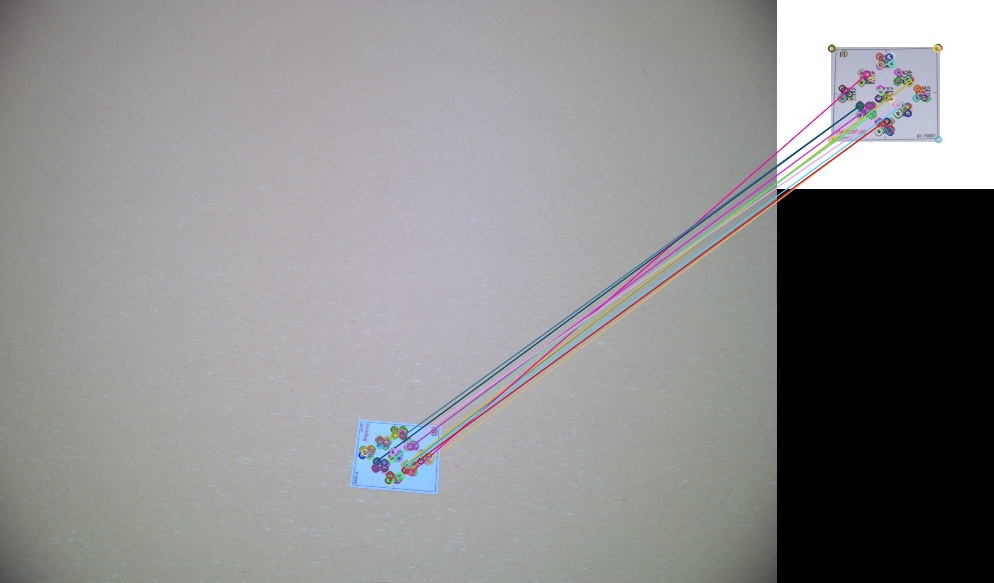}
  \caption{Feature matching to detect nodes on the warehouse ground. The large image on the left is the live image from the camera. The image on the right with the white background is the reference image from which features are computed at start.}
  \label{fig:matching}
\end{figure}

The detection of the nodes in the image is illustrated in Fig.~\ref{fig:matching}.
The cropping of a region around the nodes aims at reducing the computing time when attempting to read the DataMatrix on the nodes with the full image and when precisely detecting the position of the node in the image.
It is implemented as ORB feature matching\cite{Rublee11a}, where reference ORB features are computed at start from a single node image and then matched with the current image.
An ORB feature describes a small salient part of the image, generally at places corresponding to corners in terms of pixel intensity. The ORB feature has the double advantage over other features such as SURF and SIFT, that they are orientation invariant and patent-free.
The rotation invariance is particularly important because the camera can have all positions around the node, meaning that the node projection in the image plane can have any arbitrary orientation.

In order to isolate two or more nodes in the image, a K-means\cite{MacQueen67a} algorithm is used to separate clouds of matching features in the image.
The maximal number of clusters is set to three at the start of the algorithm and two clusters too close to each other are then merged into one because they effectively both contain features from the same node.
Due to the large distance between ground nodes and the relatively long focal distance of the objective, there is no requirement to have more clusters in the image and a simple clustering method suffices.
The position of the cluster with the highest number of features determines the center of the region of interest (ROI) around a node but the ROI is further processed only if it contains a minimal configurable number of features.

\subsection{DataMatrix decoding}

The cropped image is then sent to the algorithm in charge of reading the DataMatrix.
The chosen library for decoding DataMatrix is libdmtx\cite{libdmtx}.
In order to reduce the computing time for this part of the process, the DataMatrix are not read in the case of a blurred image and a timeout is set for the reading.
In order to determine if an image is blurred, we compute the Laplacian of the image.
For a function $f(x, y): \reals^2 \mapsto \reals$ the Laplacian is given by
\begin{equation}
  L_{f} = \frac{\partial^2 f}{{\partial x}^2} + \frac{\partial^2 f}{{\partial y}^2}
\end{equation}

For a discrete image, the Laplacian is computed by using the discrete intensity gradient, i.e.\ computed with the Sobel operator. The focus measure is then given by:
\begin{equation}
  \mathrm{focus\_measure} = {\sigma_L} ^2,
\end{equation}
where $\sigma_L$ is the standard deviation of the Laplacian values.

The threshold $t_f$ to determine whether the image is considered to be blurred depends on the mean intensity values over the complete image, $\hat{I}$, and is given by:
\begin{equation}
  t_f = \alpha \hat{I},
\end{equation}
where $\alpha$ is a scalar parameter, currently set to 0.2, that can be tuned to change how many images pass the filter.
An image is considered to be sharp when $\mathrm{focus\_measure} > t_f$.

Alternatively, to avoid motion blur, a gimbal system could have been used but this solution was then discarded because, first, a lighter device is preferred for wearable devices, secondly, the marker-based localization is a complement to the main localization algorithm, the visual odometry, and, third because, an alternative identification method was implemented, as explained latter in this paper.

The message coded in the DataMatrix codes is composed by the node identifier and a code corresponding to its position on the node itself.
The knowledge of the node identifier allows one to obtain the absolute node position in the warehouse.

\subsection{Computation of the camera position}

\begin{figure}[ht]
  \centering
  \includegraphics[width=10cm]{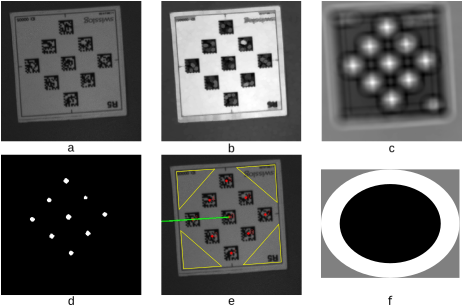}
  \caption{a - Input image; b - After opening; c - After correlation; d - After thresholding; e - Final result with determination of the orientation of the node in the image; f - Double kernel for the DataMatrix detection. The gray levels are scaled for visibility, original values are $0.5$ for black, $-0.5$ for white and $0$ for gray.}
  \label{fig:correlation}
\end{figure}

The next step in the algorithm for ground-node-based human localization is the precise determination of the position in the image of known points of the real world.
This is achieved through correlation.
Here we take advantage of the reduced image size thanks to the previous extraction of the region of interest.
The different steps of the correlation-based algorithm are:

\begin{myitemize}
  \item Resize to further reduce the necessary computing power
  \item Morphological opening to darken the DataMatrix codes (Fig.~\ref{fig:correlation}, b).
    The opening operation is run three times.
    The result of the opening operations is that the DataMatrix constituted of both black and white pixels will constitute of mostly black or dark pixels, thus allowing further localization with correlation.
  \item Intensity scale changing (black = $-$0.5, white = 0.5), so that the correlation be analog to an \emph{``exclusive or''} operation.
  \item Correlation with ``double kernel'' to detect a dark zone surrounded by a white zone.
    The size of the kernel is the expected size of the DataMatrix codes in the image (Figs.~\ref{fig:correlation}, c and~f).
    The best kernel size depends on the relative position of the node and the camera and it is chosen so that it gives the best result for a node close to the human (node projected to the bottom of the image) as that will be the most accurate for the Perspective-n-Point algorithm used further.
  The correlation operation is used to detect a black zone surrounded by a white border, i.e. the DataMatrix codes after the opening operation.
  It must be noted that the image in Fig.~\ref{fig:correlation}, c, has been rescaled for representation but does not need to be rescaled for the algorithm itself.
  The operation is costly because the usual kernel size is between 50 and 70 pixels.
  This is the reason why that correlation operation is not done on the whole image but on a cropped version around the node.
  A kernel with a ``square shape'' cannot be used because the orientation of the node projection in the image is unknown.
  An elliptic kernel is the best approximation of the shape of the DataMatrix distorted by the perspective transformation.
  \item Thresholding with a factor proportional to the maximum of the previous step to keep only the parts of the image that match the kernel (Fig.~\ref{fig:correlation}, d).
  \item Grid circle detection to detect the 3$\times$3 blob pattern in the image thanks to OpenCV's \texttt{findCirclesGrid}\cite{opencv_library}.
    The issue with this function is that the order of the circle center is unstable.
  \item Orientation detection with the detection of the node corner zone with the lowest standard deviation of pixel intensities, allowing to remove the rotation symmetry by providing 8 points instead of 9 to the next step of the algorithm.
    The zones for the orientation detection are shown Fig.~\ref{fig:correlation}, e.
    In the case that the DataMatrix can be decoded, the decoding process also provides the DataMatrix orientation in the image and this piece of information is used to compensate for any failure in the detection with corner zones because the former is more reliable.
    Determining the orientation by reading a second ground node in the image is not practical because the chances of seeing two nodes at the same time is very low, given the long objective used.
  \item Computation of the camera position relative to the node with OpenCV's \texttt{solvePnp} function.
\end{myitemize}

\section{RESULTS}

\begin{figure}[ht]
  \begin{minipage}{8cm}
    \centering
    \includegraphics[width=6cm]{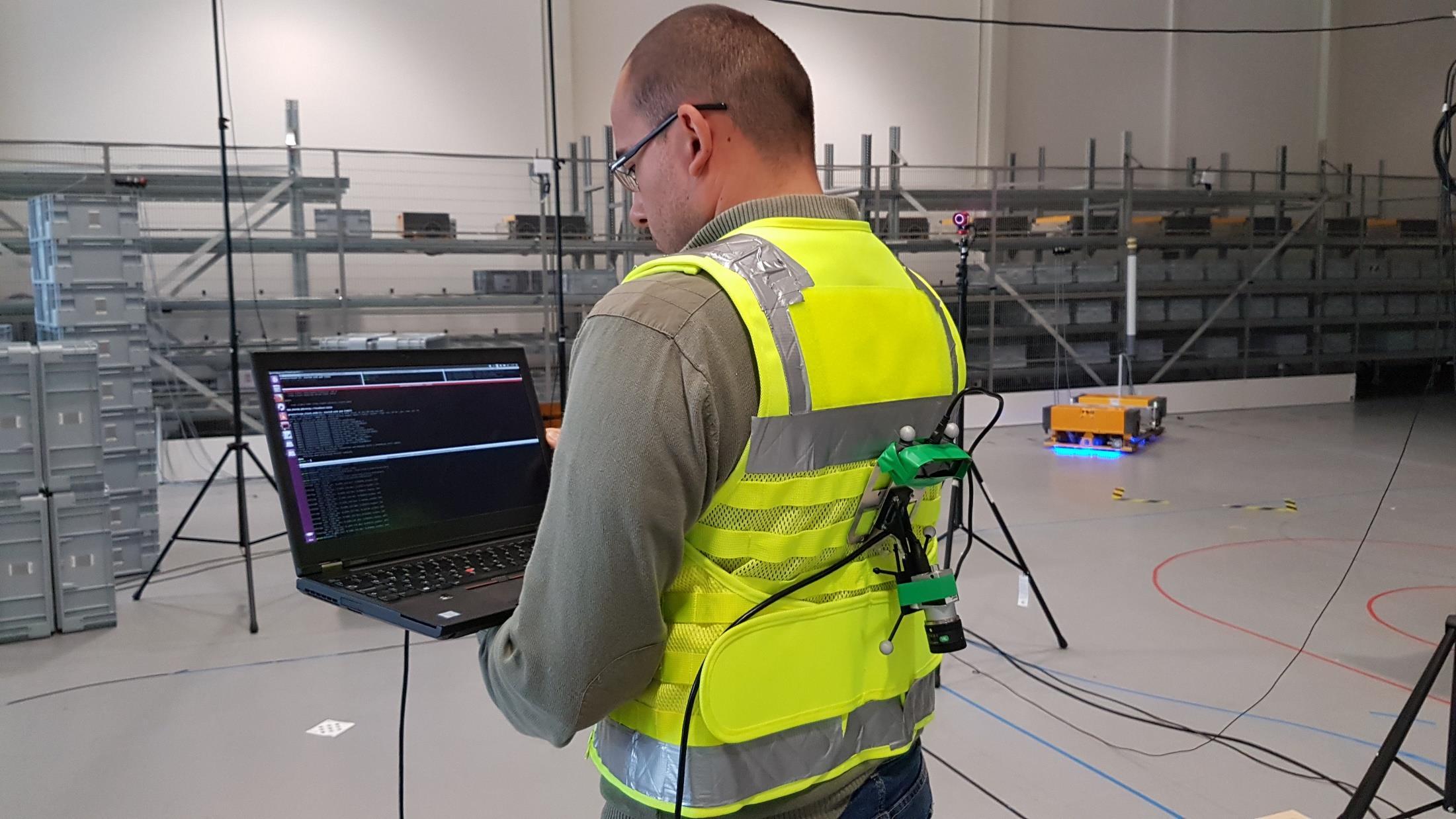}
    \caption{Human operator wearing the Safety Vest equipped with vision sensor for localization.}
    \label{fig:human_localization_device}
  \end{minipage}
  \qquad
  \begin{minipage}{8cm}
    \centering
    \includegraphics[width=6cm]{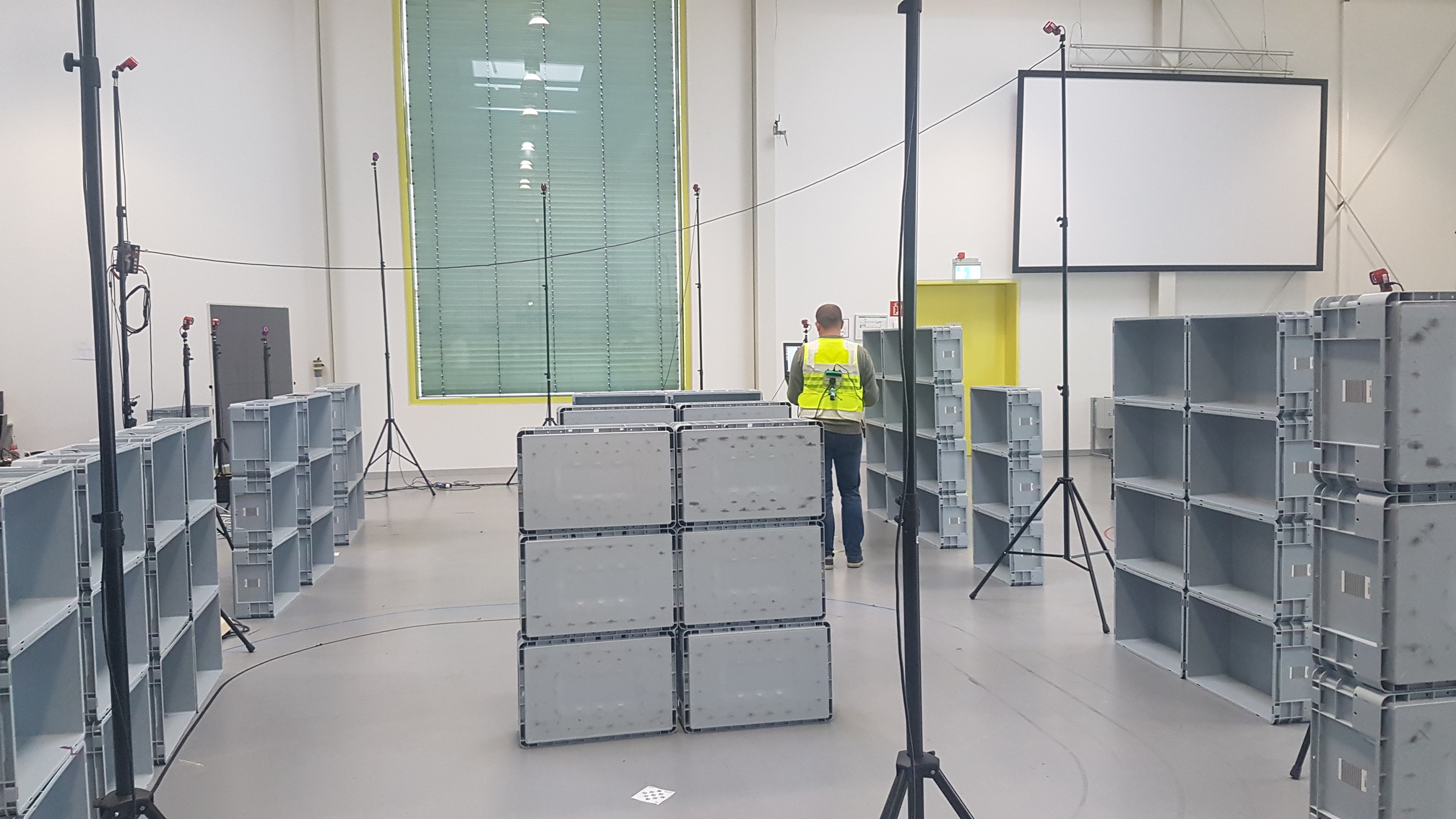}
    \caption{Simulated warehouse at the IML facilities.}
    \label{fig:dortmund_hall}
  \end{minipage}
\end{figure}

The current prototype of the human-localization device can be seen in Fig.~\ref{fig:human_localization_device}.
The camera setup design consists of a stereo camera pointing horizontally for the visual odometry and a monocular camera pointing downwards for the ground-node-based localization.
The spheres are the markers for the ground-truth localization system.

The ground-node-based localization algorithm was tested in the facilities of the Czech Technical University in Prague with ground-truth data from a Vicon tracking system.
The position of the markers on the portable device is known in the camera reference frame, so that the position given by the algorithm can be directly compared to the ground truth.
The tests presented here were carried out by moving the camera by hand and the movements are therefore smoother than when carried on the back as was done during the recording of a prior dataset at the Fraunhofer IML facilities, Dortmund, Germany, one of the partners in the SafeLog Project.
The algorithm was implemented in ROS\cite{Quigley09}, the Robot Operation System, a node-based system with standard messages that allow easy collaboration with other project partners.
The mean processing time of one frame is \unit[130]{ms}, including image rectification and multiple image serializations/deserializations beetween ROS nodes on an Intel i7 8$^\mathrm{th}$ Gen processor.

\begin{figure}[ht]
  \begin{minipage}{0.475\columnwidth}
    \centering
    \includegraphics[width=6cm]{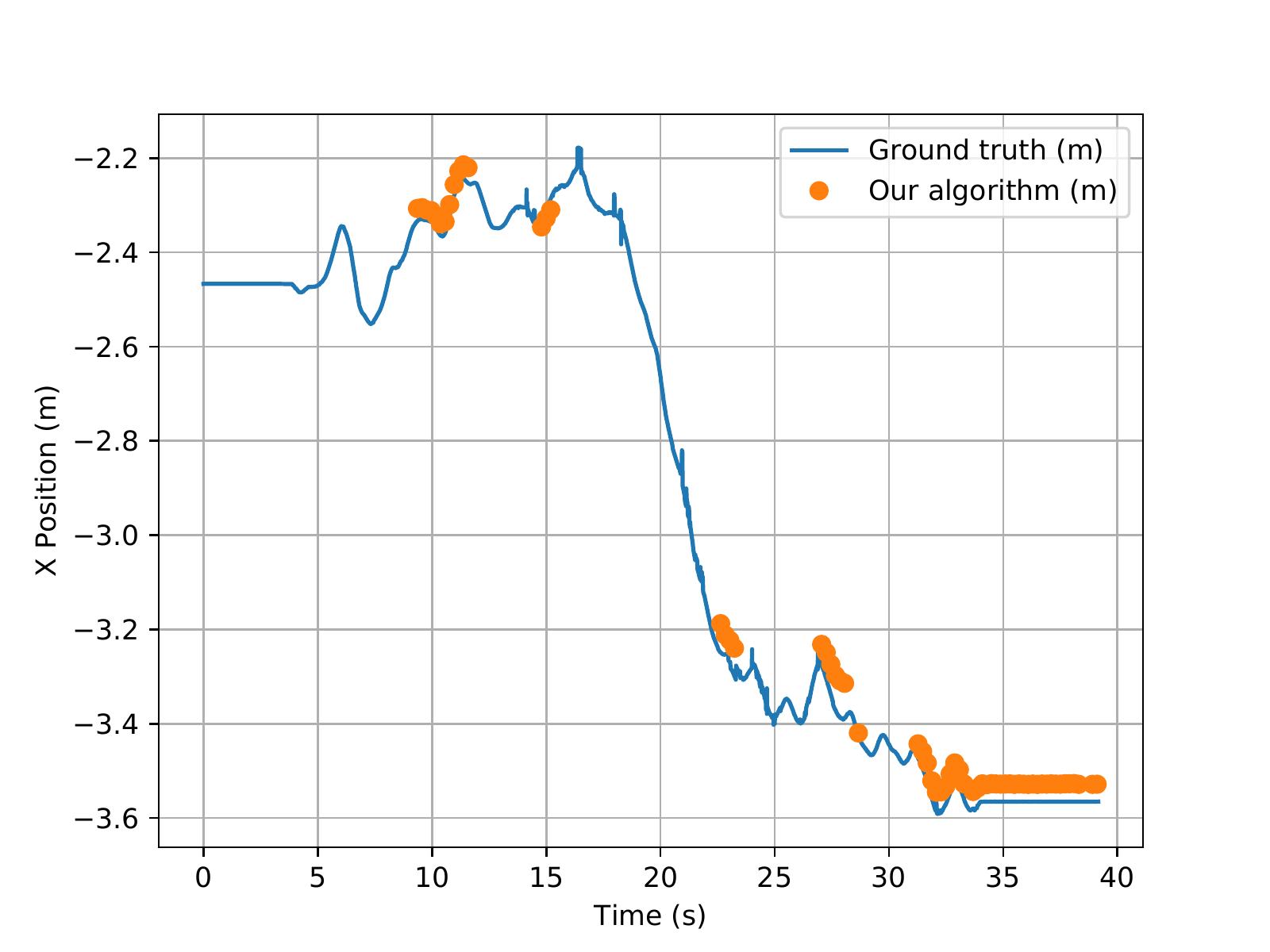}
    \caption{Ground-node-based localization versus ground-truth in the x-direction. The x-axis is the time in seconds, the y-axis is the position in the world coordinate system in meters.}
    \label{fig:2018-12-05--09-22-position-x}
  \end{minipage}
  \qquad
  \begin{minipage}{0.475\columnwidth}
    \centering
    \includegraphics[width=6cm]{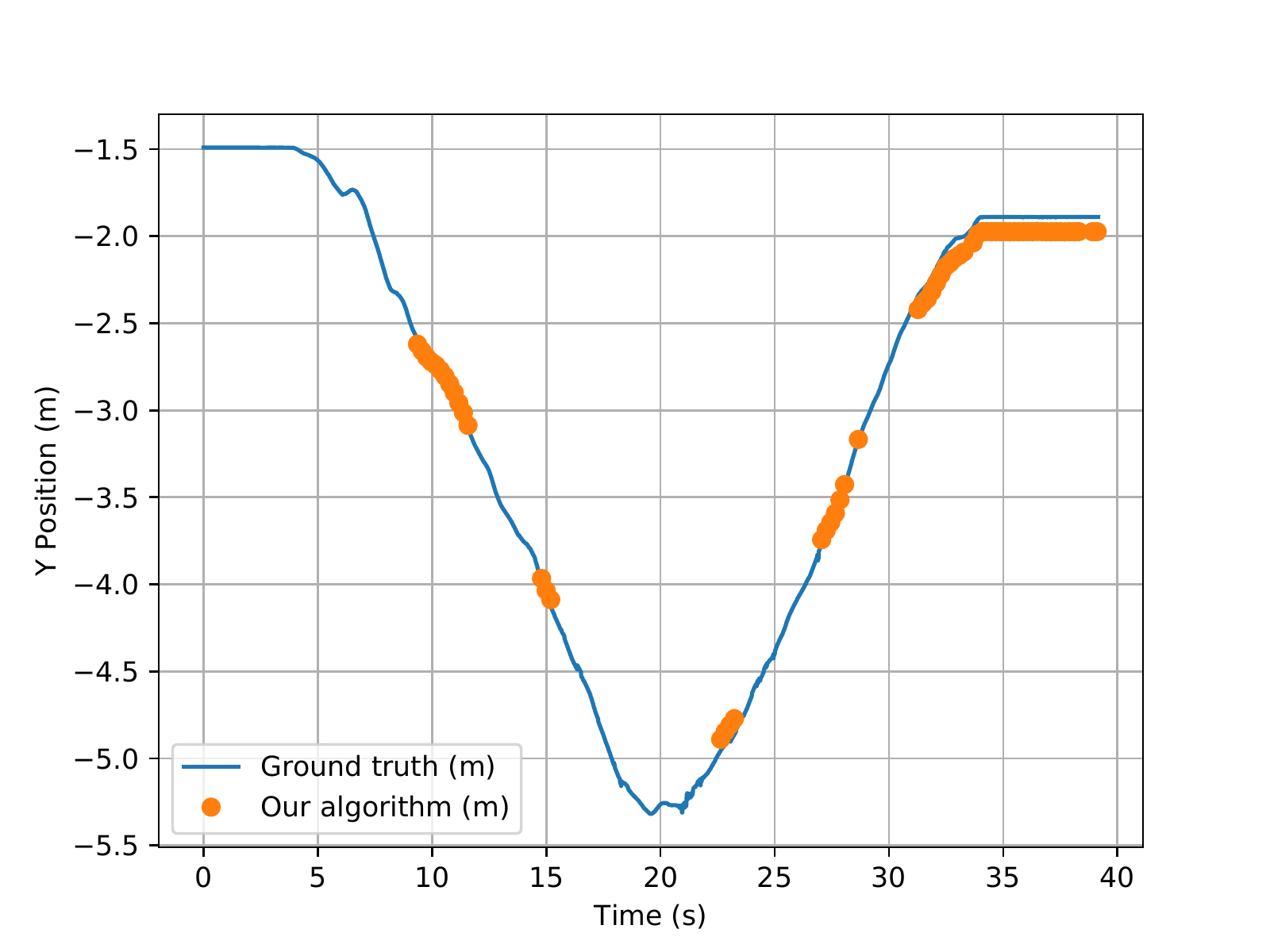}
    \caption{Ground-node-based localization versus ground-truth in the y-direction. The x-axis is the time in seconds, the y-axis is the position in the world coordinate system in meters.}
    \label{fig:2018-12-05--09-22-position-y}
  \end{minipage}
\end{figure}

The results of the localization system are shown in Figs.~\ref{fig:2018-12-05--09-22-position-x},~\ref{fig:2018-12-05--09-22-position-y}.
The long intervals between values given by the localization system is not an issue because the ground-node-based localization is only part of a complete localization system, the visual odometry providing a pose at regular intervals even is both the monocular camera and the stereo camera are obstructed.

\begin{figure}[ht]
  \begin{minipage}{0.475\columnwidth}
    \centering
    \includegraphics[width=6cm]{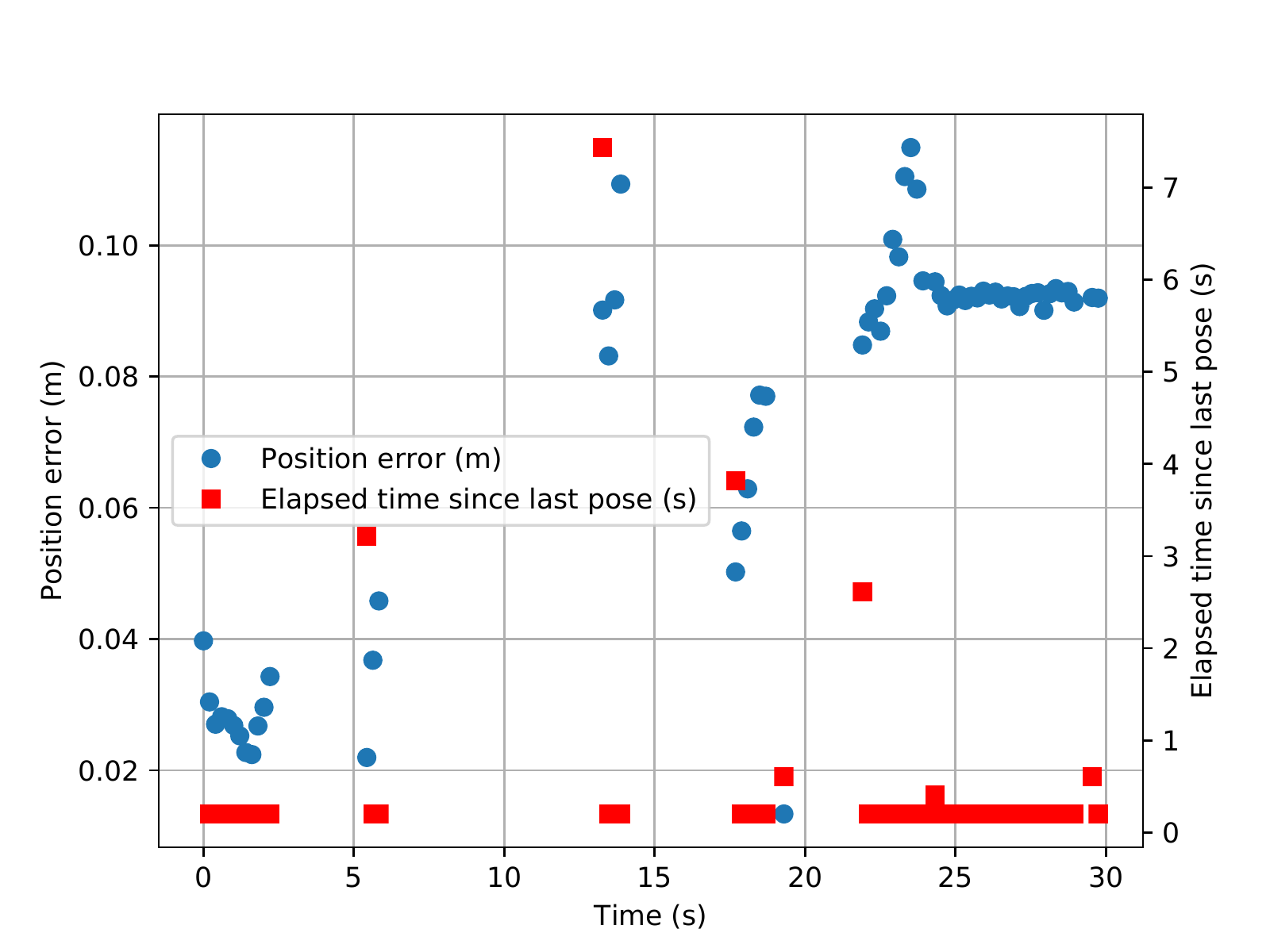}
    \caption{Position error of the ground-node-based localization system.}
    \label{fig:2018-12-05--09-22-position_error_elapsed_time}
  \end{minipage}
  \qquad
  \begin{minipage}{0.475\columnwidth}
    \centering
    \includegraphics[width=6cm]{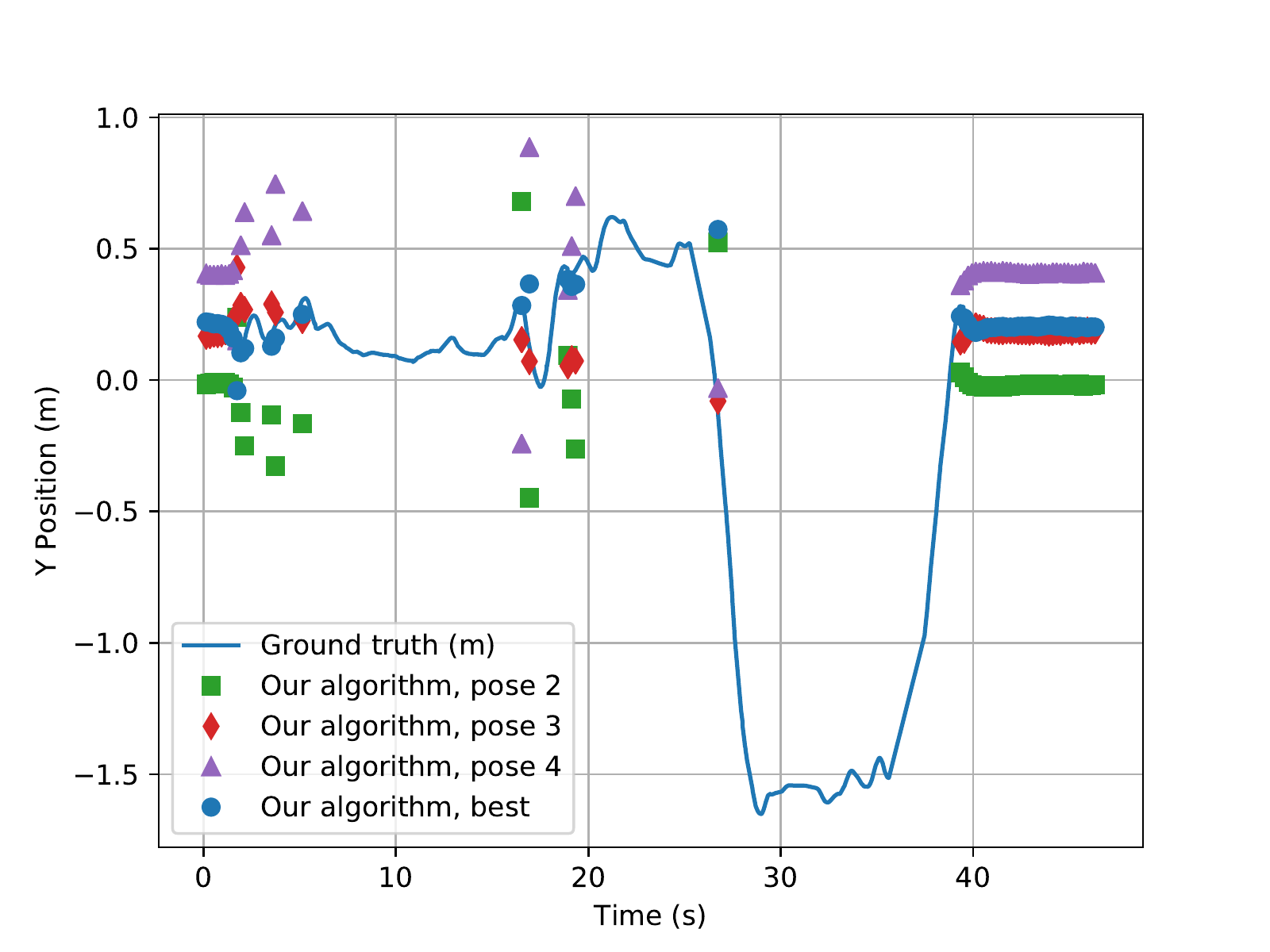}
    \caption{Ground-node-based localization versus ground-truth in the y-direction for a dataset taken in the IML facilities. The x-axis is the time in s, the y-axis is the position in the world coordinate system.}
    \label{fig:2018-09-07--14-38-position-y}
  \end{minipage}
\end{figure}

Fig.~\ref{fig:2018-12-05--09-22-position_error_elapsed_time} shows the position error ($\sqrt{\mathrm{d}x^2 + \mathrm{d}y^2}$) for the localization system and the elapsed time since the last pose determination was done.
It can be seen that the localization error does not exceed \unit[12]{cm}.
In the case that a ground node can be seen in the image, a new pose is computed every \unit[0.2]{s}, what corresponds to the period of arrival of a new input image.
The longer periods where no pose could be computed are due to the fact that the camera does not see any entire ground node.

Another prior dataset was recorded in the IML facilities in Dortmund with a setup closer to the real application.
The person who recorded the dataset wore a safety vest with a camera setup on his back, cf. Fig.~\ref{fig:human_localization_device}.
In this dataset the ground nodes were placed across the arena.
We used five nodes and for each node we measured their exact location in the global reference frame.
Due to the lack of real racks, we have built the walls and racks out of boxes but this has little influence on the ground-node-based localization algorithm, as long as they form alleys, cf. Fig.~\ref{fig:dortmund_hall}.
The arena was equipped with the OptiTrack system.
We placed OptiTrack makers on the camera setup, which was then tracked by the OptiTrack system during the dataset recording as ground truth.

We tested various scenarios which we expect to appear in the warehouses:
\begin{myitemize}
  \item fast or slow walk through halls,
  \item forward or backward walk,
  \item standing or crouching between the racks.
\end{myitemize}

All the recordings begin with a person standing above a ground node, so that we can compute the position in the global reference frame, what is important for the visual odometry.

Fig.~\ref{fig:2018-09-07--14-38-position-y} shows the result of the ground-node-based localization for one of the most challenging datasets from those taken in the IML facilities.
In this dataset, the device wearer alternated between fast walking and crouching between rows of boxes.
The light during the recording of this dataset was as low as 160 lux.

\begin{figure}[ht]
  \begin{minipage}{8cm}
    \centering
    \includegraphics[width=6cm]{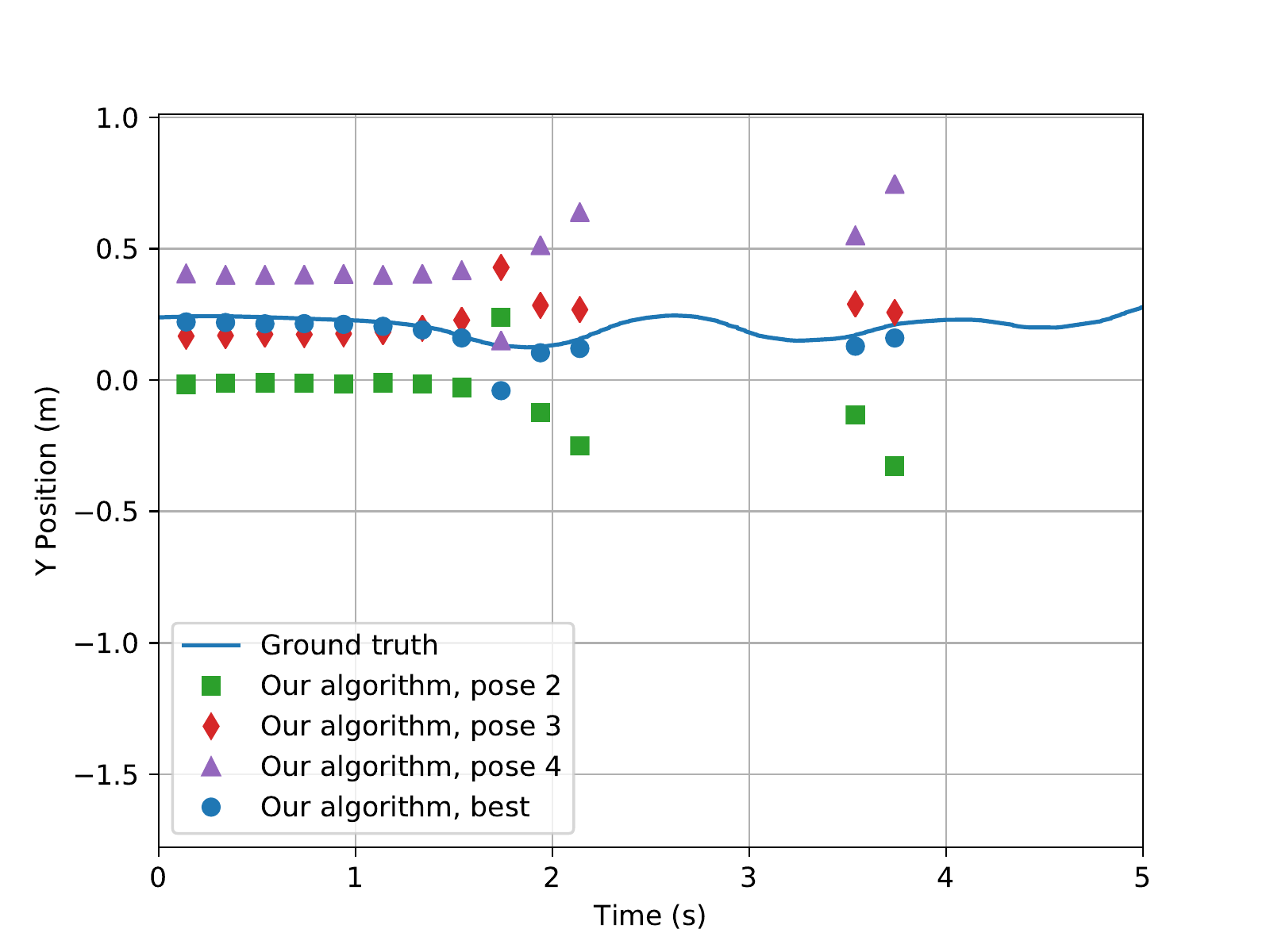}
    \caption{Four poses versus ground truth for the IML dataset.}
    \label{fig:2018-09-07--14-38-four_poses-y-zoom}
  \end{minipage}
  \qquad
  \begin{minipage}{8cm}
    \centering
    \includegraphics[width=6cm]{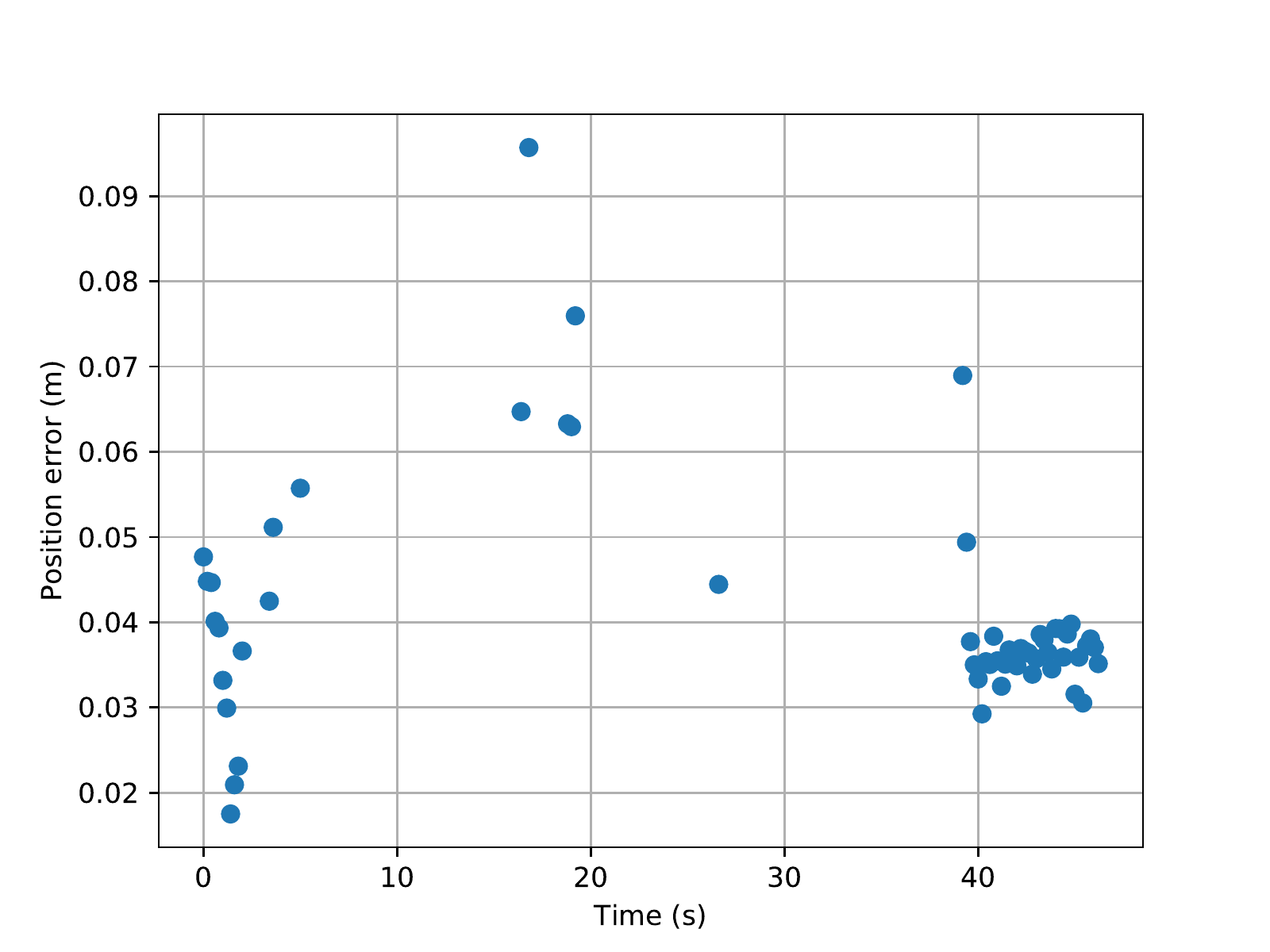}
    \caption{Position error of the filtered result of the ground-node-based localization.}
    \label{fig:2018-09-07--14-38-position_error-filtered}
  \end{minipage}
\end{figure}

The ground-node-based localization algorithm actually provides the four poses corresponding to the four possible node rotations.
The first pose will be the one the algorithm thinks is the correct one but it also happens that the rotation cannot be determined, for example when the  standard deviation of the intensity in the four corner differs less than a given threshold and the DataMatrix cannot be decoded.
In this case, the pose order is arbitrary.
This effect can be seen in Fig.~\ref{fig:2018-09-07--14-38-four_poses-y-zoom} that shows a zoomed zone of Fig.~\ref{fig:2018-09-07--14-38-position-y} with the four poses computed by the algorithm.
It can be clearly seen that at time \unit[1.7]{s} the pose 4 is the correct pose, not the one the algorithm chose as correct.

In order to counter-balance this phenomenon, the pose can be filtered with respect to the pose given by the visual odometry running in parallel.
In the results presented Fig.~\ref{fig:2018-09-07--14-38-position_error-filtered}, the pose from visual odometry is replaced with the ground-truth pose.
The visual odometry being a relative localization system, the need to compute a pose without external output to obtain an absolute localization is important.
In the warehouse this will be solved by initializing the localization system at a known and stable position so that it can be ensured which of the four computed poses is the correct one.
The position error of the filtered data is show in Fig.~\ref{fig:2018-09-07--14-38-position_error-filtered}.
The maximal localization error in this trial is below \unit[10]{cm} and the algorithm is able to estimate which of the four poses is the correct one with a success rate of \unit[95]{\%}.

\section{ALTERNATIVE IDENTIFICATION}

The issue with the current workflow is that in the case that the node cannot be identified the camera position cannot be computed because the absolute position of the node in the warehouse is unknown.
The identification of a node relies on the decoding of one of the DataMatrix codes that is rendered difficult by their small size of \unit[15]{mm}, the long distance to the camera and motion blur.
Using larger tags or other tags such as AprilTags\cite{Wang16a} would have been beneficial for the localization but the robot localization of the warehouse system already relies on the current ones and this cannot be changed within the scope of the SafeLog project.
As already mentioned the ground-node-based localization algorithm is not meant to be standalone but is complemented by a visual odometry algorithm, which has a high output rate but is relative and has a drift.
We exploit the pose given by this combined localization algorithm to back project the detected node on the warehouse floor and identify this node as being the closest to all nodes in the node database with the condition that the last absolute detection was not too long ago.

The position of the projection of the node's center on the image plane after image undistortion is given by $C_p = (c_{px}, c_{py})^\mathrm{T}$, in pixels.

Let us project this point on the plane $z = \unit[1]{world\_unit}$ in the camera frame through the projection around the camera center. The coordinates of this point, $C_c$ are
\begin{equation}
  C_c = \begin{pmatrix}
    \frac{\left(c_{px} - c_x^\prime\right)}{f_x^\prime} \\
    \frac{\left(c_{py} - c_y^\prime\right)}{f_y^\prime} \\
    1
  \end{pmatrix}_{\mathrm{R}_c},
\end{equation}
where $f_x^\prime$ and $f_y^\prime$ are the focal lengths in $x$ and $y$ directions, $c_x^\prime$ and $c_x^\prime$ are the coordinates of the image center in pixel coordinates, and $\mathrm{R}_c$ is the camera reference frame.

The coordinates of this point in the world reference frame is given by
\begin{equation}
  C_w = {}^w\mathrm{T}_c \begin{pmatrix}
    C_c \\ 1
  \end{pmatrix},
\end{equation}
where ${}^w\mathrm{T}_c$ is the homogeneous transform from camera frame to world frame.

The assumption is made that the floor plane has its normal along the $z$-axis of the world reference frame, i.e.\ that it is defined by the equation $z = h$, where $h$ is the floor height.
The ray passing through the camera center and point $C_w$ intersects with the plane representing the floor at position $C_f$ such that $C_f = C_0 + t\left(C_0 - C_w\right)$ and $C_{0,z} = h$, with $t \in \reals$.
This gives
\begin{equation}
  C_f = \begin{pmatrix}
    C_{0,x} + t \left(C_{0,x} - C_{w,x} \right) \\
    C_{0,y} + t \left(C_{0,y} - C_{w,y} \right) \\
    h
  \end{pmatrix},
\end{equation}
with $t = \frac{h - C_{0, z}}{C_{0,z} - C_{w,z}}$, where $C_{0,\cdot}$ are the coordinates of the camera optical center in the world reference frame.
The camera cannot be in the floor plane so that $C_{0,z} \neq C_{w,z}$.

\begin{figure}[ht]
  \centering
  \includegraphics[width=6cm]{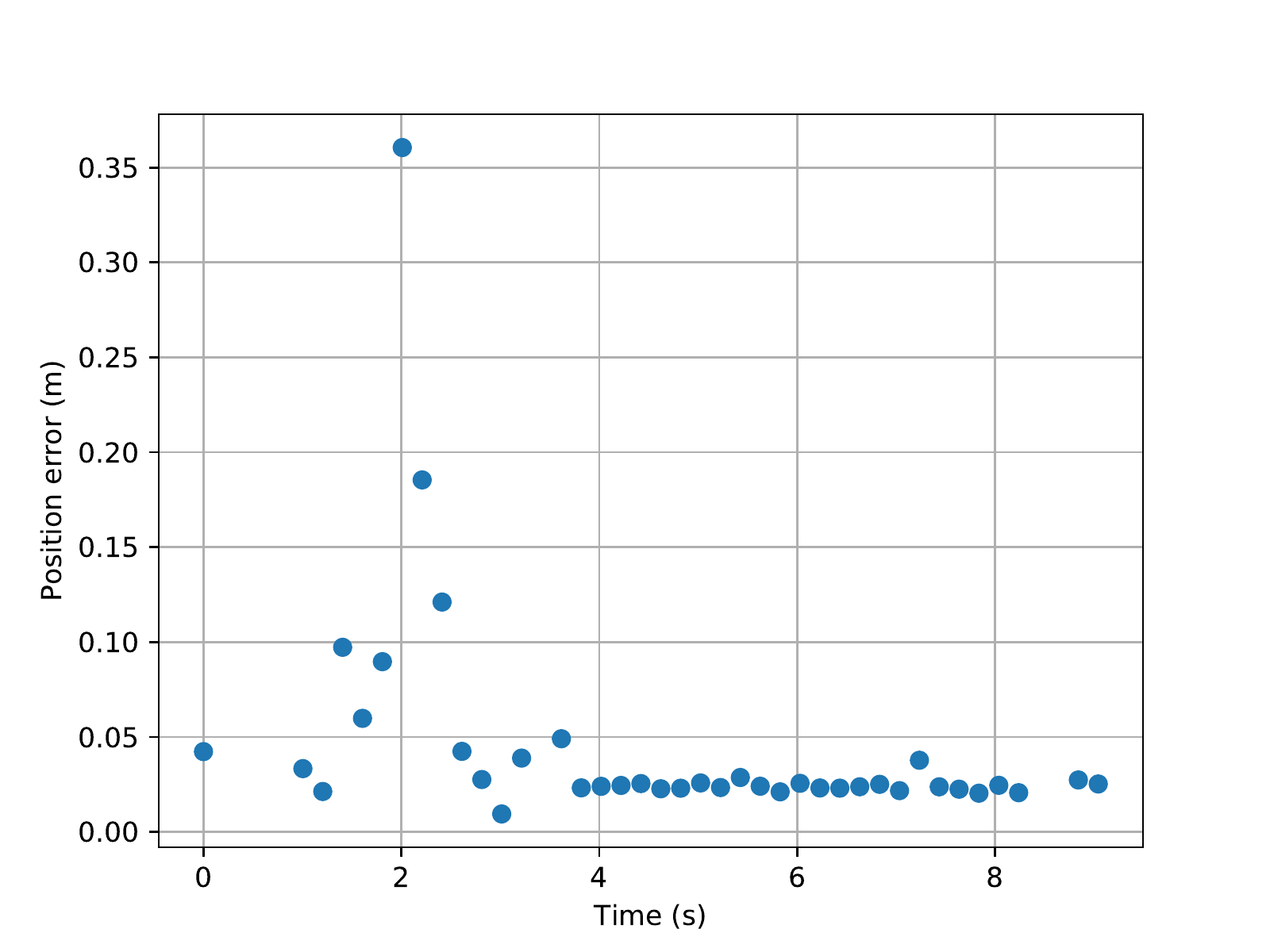}
  \caption{Position error between the projection of the node image on the floor and the closest node in the database}
  \label{fig:stickers_projector-position_error}
\end{figure}

The results of the node projection algorithm is presented in Fig.~\ref{fig:stickers_projector-position_error}.
They show the position error between the projection of the node image on the floor and the closest node in the database for the dataset used in Fig.~\ref{fig:2018-12-05--09-22-position-x} to~\ref{fig:2018-12-05--09-22-position_error_elapsed_time}.
The projection error is mostly of a few centimeters and under \unit[40]{cm} so that it can be used to identify the node detected in the image.

\section{CONCLUSION}

We presented an absolute localization system for a human operator in a warehouse that uses artificial markers readily available in the warehouse in question.
The algorithm provides the position despite low-light conditions, quick human movements, and long distances to the markers.
The frequency of the data depends on whether a ground node can be seen in the image or not but when this is the case the algorithm provides an full pose at at least \unit[5]{hz} with an accuracy below \unit[20]{cm}.

Future work will consists in the integration of the ground-node-based localization with the visual odometry algorithm for a complete localization system as well as further testing.

\acknowledgments 

This work is supported by the SafeLog project funded by the European Union's Horizon 2020 Research and Innovation Programme under grant agreement No.~688117 and by the European Regional Development Fund under the project Robotics for Industry 4.0 (reg. no. CZ.02.1.01/0.0/0.0/15 003/0000470).

\bibliography{bibliography} 
\bibliographystyle{spiebib} 

\end{document}